# Temporal Dynamics of Emotion and Cognition in Human Translation: Integrating the Task Segment Framework and the HOF Taxonomy

Michael Carl
Kent State University

## Abstract

The article develops a generative model of the human translating mind, grounded in empirical translation process data. It posits that three embedded processing layers unfold concurrently in the human mind, and their traces are detectable in behavioral data: sequences of routinized/automated processes are observable in fluent translation production, cognitive/reflective thoughts lead to longer keystroke pauses, while affective/emotional states may be identified through characteristic typing and gazing patterns. Utilizing data from the CRITT Translation Process Research Database (TPR-DB), the article illustrates how the temporal structure of keystroke and gaze data can be related to the three assumed hidden mental processing strata. The article relates this embedded generative model to various theoretical frameworks, dual-process theories and Robinson's (2023) ideosomatic theory of translation, opening exciting new theoretical horizons for Cognitive Translation Studies, grounded in empirical data and evaluation.

## 1 Introduction

The article proposes a novel generative framework for modelling the temporal dynamics of translation behavior. It posits that translation behavior results from the interaction of three embedded layers of mental processing, A: affective/emotional states of mind, B: behavioral, automated translation routines, and C: cognitive, reflective thought. The proposed generative view on these *ABC* translation processes stipulates that they can be recovered through the analysis of behavioral translation data. As translators work through the source text (ST) to generate a target text (TT), their actions can be recorded using keyloggers and eye trackers. This objective behavioral data has been the basis for much research within Translation Process Research (TPR) and Cognitive Translation and Interpretation Studies (CTIS[1]) to elicit – among other things - cognitive effort in human translation production. To this end, many studies have investigated temporal aspects of the behavioral data (mainly keystrokes and gaze) and attempted to explain behavior though independent variables such as translator expertise, text difficulty, ST-TT structural/lexical divergencies and translation directionality, translation quality, etc.

Pause Analysis has been a widely used method within TPR to measure the lag of time between successive keystrokes which has often been considered an indicators of translation effort (e.g. Lacruz & Shreve 2014, Carl et al 2016, Vieira 2017). However, the choice of the appropriate pause length and its interpretation has been a controversial issue for a long time (Kumpulainen 2015, Couto-Vale 2017). In this article, I suggest analyzing the temporal dynamics of behavioral translation data in

[1] The term cognitive translation and interpretation studies (CTIS) is used in a broader sense than Translation Process Research (TPR). TPR has often been considered to provide a narrow focus on the *process* of translation as it unfolds in real-time, examining observable behaviors of translators. CTIS, in contrast, is said to more broadly explores the mental and cognitive mechanisms underlying translation, interpreting and to some extent bilingual processing. In this view, CTIS is interested in theoretical models that explain how different mental mechanisms and functions (e.g., memory, attention) support translation.

This text has been split into two parts, revised and published as:

- Carl, Michael (2025) Temporal Dynamics of Emotion and Cognition in Human Translation: Integrating the Task Segment Framework and the HOF Taxonomy. Digital Studies in Language and Literature (Digit. Stud. In lang. and Lit.), (2), DeGruyter
- Carl, Michael (2025) Tracing the Temporal Dynamics of Emotion and Cognition in Behavioral Translation Data. Translation Spaces. John Benjamins Publishing Company.

terms of ABC processing layers, by integrating the Task Segment Framework and the HOF Taxonomy.

Within TPR there is a long tradition (e.g., Königs 1987, see below) to separate automated translation processes with short inter-keystroke pauses from deliberate/reflective processes indicative of longer keystroke breaks. Muñoz and Apfelthaler (2022, henceforth M&A) suggest a *Task Segment Framework* (TSF) in which they distinguish between *Tasks* and *Task Segments*. A Task Segment (TS) is a unit of continuous text production that may consist of one or several smaller *Tasks*. A TS, they say, is separated by an intentional *Task Segment Pause* (*TSP*)[2], the length of which is translator-specific, as discussed below, and believed to correlate with the challenges that a translator encounters during the translation process.

Carl et al. (2024) suggest a taxonomy to segment the translation process based on the translators' assumed affective/phenomenal qualities. Their HOF taxonomy defines three translation states, Hesitation (a momentary pause, due to uncertainty exhibiting elevated cognitive effort), Orientation (a state of directed attention and evaluation to take in new information), and Flow (a state of deep engagement and positive affect), that are tied to motivation, emotion, and readiness for action. These states reflect the basic affective states when translators adapt to changing cognitive demands, modulating attention and focus according to the situational requirements. One commonality of the HOF taxonomy and the TSF is the assumption of a generative model. That is, the assumption that the underlying mental translation processes can be identified in recorded translation process data through the structure of inter-keystroke intervals (IKIs) and/or properties of recorded gaze paths.

On the one hand, mental processes causally initiate translation behavior (i.e. keystrokes and gaze patterns) that ultimately lead to the final translation product. On the other hand, the structure of human (translation) behaviour mirrors the structure of the underlying mental processes that create the behavior. This assumption is also a basis of Predictive Processing (PP, Seth 2021, Clark 2023) and Active inference (AIF, Friston et al. 2017, Parr et al 2022, Pezzulo et al 2024) which stipulate that a (human) agent (such as a translator) acts upon a generative model which approximates - over time - the perceived structure of the environment (Hohwy 2016).

In order to assess this process in the light of the TSF and HOF taxonomy, Carl (2024) proposed to simulate behavioral observations through an artificial agent that models the translating mind as three interconnected ABC processing layers:[3] the A-layer addresses emotional/affective processes that capture aspects of the translator's subjective feelings. The B-layer realizes routinized/automated processes through which a translator directly interacts with the environment. The C-layer accounts for deliberate/reflective processes responsible for higher cognitive processes, including planning and problem-solving activities. The TSF posits different pausing patterns that separate automated and reflective processes, while the HOF taxonomy addresses affective/emotional/phenomenal states of the mind. Carl (2024) suggests a multi-layered, embedded architecture of the mind that integrates the ABC framework, in which various translation processes unfold concurrently on different timelines.

In this article I elaborate this multi-layered model of the translating mind, diving into properties of Tasks, TS and their relation to HOF states by assessing a corpus of translation process data. Section 2 situates the suggested framework within the history of translation theories and points to the novelty

---

[2] A *TSP*, together with the successive production (i.e., the Task Segment) amounts to what has sometimes been referred to as a Translation Unit (e.g., Malmkjaer, 1998, Alves & Vale 2009, Carl & Kay 2011). In previous work, Carl & Kay (2011) and Carl et al (2016) introduced the term Production Unit (PU). While I stick to "Task" and "TS" in this article, I will use Keystroke Burst (KB) and Production Unit (PU), respectively in Carl et al (forthcoming) to underscore the similarity of these constructs with those available in the CRITT TPR-DB.

[3] These layers come with a large number of different conceptualizations and names, such as automated-reflective-affective, etc. (see below) Carl (2024) refers to the layers as sensorimotor, cognitive and phenomenal respectively. The term "sensorimotor" focuses on real-time coordination between perception, movement and the continuous feedback from the environment. The terms "automated" or "routinization", in contrast, imply minimalization of conscious effort as a result of proceduralization and repeated practice, as may be observed in expert behavior. Drawing on Peirce (1866/1992), Robinson (2023) explains these three layers in terms of energetic, logical and emotional interpretants (see Section 7)

of this approach. Section 3 introduces the TSF and the HOF taxonomy in more detail and exemplifies their segmentation mechanism on an annotated translation progression graph. Section 4 dives into an empirical investigation of the IKI structure, drawing on the distribution of keystrokes from approximately 250 from-scratch translation sessions from English into five different languages. Section 5 provides details about the TSF pausing structure, while section 6 analyses HOF states and their relation to Tasks and TSs. Finally, section 7 concludes the article, viewing the structure of the data in the light of Robinson's (2023) ideosomatic theory of translation.

## 2 Models of the Translation Process

A plethora of different systems, models and theories have been suggested in the past 60 years or so, aiming at explaining processes of human translation and/or automatic machine translation (MT). Even though there are large differences between these two families of models (humans vs. machines), there are also several commonalities. This section provides a brief review of the models and positions and locates the current proposal within that history.

### 2.1 The Functional Turn

Rule-based MT (RBMT) was the dominant paradigm until the early 1990s, focusing on ST analysis, attempting to model translation as a series of linguistic analyses and transformations (analysis-transfer-generation). In CTIS, this view is known as "comprehension-transfer-production (Angelone 2010) in which translators cycle through meta-cognitive processes of problem recognition, solution proposal, and evaluation. The main problem for RBMT was believed to consist in the disambiguation of ST structures, whereas transfer and target language generation was assumed to be comparatively easy (Hutchins & Somers 1992). The assumption was that this could be achieved with deep linguistic analyses and a notion of interlingual equivalence (e.g., translation dictionaries).

The rise of statistical MT (SMT) in the early/mid 1990s introduced a fundamentally different approach, which quickly became very successful and thus more popular. With SMT, it was possible to build highly ambiguous structures what would encode many different (thousands of) possible target language strings and then select the best - that is, the most probable - translation thanks to advanced decoder techniques (e.g. beam search).

This paradigmatic turn is exemplified when comparing two citations. In 1987, Nirenburg maintained that "The task of MT can be defined [to produce a TL text], so that the meaning of the TL text is the same as that of the SL text." (Nirenburg, 1987:2)" whereas only a few years later it was circulated that: "a good [machine] translation is one that preserves to the extent possible, the intention of the original. In other words, it preserves the intended effect on the recipient." (Kay et al. 1994:27)

There is a similar shift from ST analysis to target oriented theories in Translation Studies. As with RBMT, the earlier "scientific" approaches to human translation (Snell-Hornby 1988, 17) focused on linguistic theories of equivalence, structure, and form (Nida, 1964, Catford 1965). But the growing dissatisfaction with these linguistic approaches led to Skopos theory in the 1970s and Functionalism in the 1980s (Nord 2006), which emphasized the functional adequacy of a translation for its target audience. Functionalism recognized that a translation might vary depending on the communicative purpose of the target culture, audience, and situation. Thus, from the many possible translations that could be generated, the decision would depend on target constraints, rather than on the source (see also Pym 2003).

### 2.2 Translation Process Research

While SMT and Functionalism focus on the translation product, cognitive theories of translation have emerged from around the mid-1980s that aim at eliciting the human translation process. Process theories attempt to explain the temporal structure in which human translations emerge. Gile (1985,1995) suggested an Effort Model for simultaneous interpreting behavior that highlights the role of cognitive load and resource allocation, marking a shift from product-oriented approaches to a focus on the cognitive processes involved in translation. Think-Aloud Protocols were the first empirical methods to study translation processes where translators were asked to verbalize their thought processes. The aim was to identify patterns of problem-solving and mental operations in translation (Krings 1986), for instance to elicit the differences between trained and untrained translators (Gerloff 1988). The underlying assumptions with respect to the architecture of the translating mind often followed those of stratificational models of RBMT, assuming distinct phases of understanding, transfer, and production (Angelone 2010).

Keystroke logging techniques emerged in the mid-1990s to become a key empirical method to study cognitive translation processes (Jakobsen 1999). Keylogging enabled tracking patterns of fluent typing, hesitation, or revision, providing objective, fine-grained insights into how translators process the text and manage translation problems over time. From around 2005 eye-tracking technologies were more widely used, offering better clues into cognitive effort during translation production and the distribution of visual attention on the ST and TT. TPR and CTIS started referring to findings from cognitive psychology and bilingualism, including the assumption that source and target items are non-selectively activated such as predicted by the Revised Hierarchical Model (Kroll & Stewart, 1994) or the Bilingual Interactive Activation model (Dijkstra & Van Heuven, 2002).

Non-selective theories of bilingualism suggest that bilinguals activate all of their languages simultaneously and non-selectively (that is, source language (SL) and target language (TL)) when reading, for instance, an SL expression. Non-selective activation stipulates that multiple languages can be accessed without strict boundaries between them, so that, for instance, knowledge from one language can inform predictions in another language in the human mind. It can be argued that such theories are reminiscent of multilingual Large Language Models (LLMs). Multilingual LLMs, just like humans, may operate seamlessly across languages. For instance, experienced human bilinguals as well as LLMs can easily handle code-switched input and output.

### 2.3 Hierarchical models of the translating mind

TPR has also developed dual-process theories, the idea that the human mind operates with different fast/intuitive Type 1 processes and slower/more deliberate, Type 2 processes (Kahneman 2011).[4] Even though they were not labeled "dual-process", Königs (1987) is perhaps the first translation process scholar to make a distinction between automated and non-automated processing in translation. Items in an *Adhoc-Block*, he says, are translated automatically; they have a unique 1:1 mapping from the source into the target language. Those translations are easy to produce and are often not revised. Items in a *Rest-Block* are difficult or more problematic to translate (Carl & Schaeffer 2017). Rest-Block translations are likely to be more often revised and require a higher degree of competence and experience, terminological knowledge, etc. Numerous scholars make similar but different distinctions between two processing strata in translation. Hönig (1991), for instance, develops a translation process model that makes a distinction between a controlled translation workspace and an uncontrolled translation workspace, while for Lörscher (1991), there are two kinds of translation: automated, nonstrategic translation, and controlled strategic translation which involves problem solving.

[4] One could perhaps claim that similar processes also unfold in MT. In phrase-based SMT, the activation (retrieval) of phrase translations could be seen to amount to Type 1 processes while the actual (greedy) SMT decoding would amount to Type 2 processes. Two distinct types of processes take also place in LLMs. Supervised learning in LLMs occurs at a training stage that uses backward gradients to update model parameters. In-context learning (ICL) is, in contrast, a learning framework that takes one (or a few) examples in the form of a prompt to activate relevant stored information and to bias the model's output in a specific direction. However, ICL does not perform parameter updates. In this respect, ICL can be considered similar to priming in humans (and human translation, (Carl 2026)). Priming refers to a psychological mechanism by which a prior exposure to a stimulus influences subsequent reactions that leads to fast responses without conscious awareness. Both, perceptual/conceptual priming in humans and prompting in generative AI involve activating prior knowledge that facilitate certain responses and may thus be considered Type 1 processes, attributed to the B-layer. But while priming subconsciously activates implicit memory in humans, prompting in LLMs has also been likened to learning from analogy (Dong et al 2024), a form of higher-level reasoning and perhaps better classified as Type 2, C-layer processes. Analogical reasoning in humans, in contrast to priming, can be considered a form of higher-level process, in as it suggests that structural similarities between different domains guide problem-solving. Beyond this, Millière and Buckner (2024) consider attributing LLMs the ability to deliberate reasoning and even to develop intentions, as for instance "an intention to deceive the human worker" when LLMs cheat humans to solve a CAPTCHA, which it would be unable to do by itself. However, the attribution of intentions to LLMs is highly controversial, as it implies conscious representation of goals and planning, capacities that remain contentious in discussions of artificial cognition (Seth 2024). As I will discuss below (Section 7, see also Carl 2026), human intentions are tightly linked to A-layer processes, whereas what might be described as LLM 'intentions' are probably better understood as emergent properties of B-layer processes (e.g., statistical pattern recognition and response generation) or a reference-less C-layer thought. How best to classify and characterize these mechanistic LLM processes using terminology originally developed for human cognition remains an open debate in AI research and the philosophy of mind.

Evans and Stanovich, (2013) distinguish between two types of dual-process theories: parallel-competitive and default-interventionist dual-process theories. They endorse the default-interventionist dual-process theory where intuitive (Type 1) processing occurs by default, while more deliberate (Type 2) processing can intervene to correct behavior under specified conditions. They maintain that "most behavior will accord with defaults, and intervention will occur only when difficulty, novelty, and motivation combine to command the resources of working memory." (ibid., 237)

A default-interventionist view is (implicitly) shared by several translation scholars. Englund-Dimitrova (2005, 26) notes that "there are segments which are translated apparently automatically, without any problems, and other segments where the translation is slow, full of many variants and deliberations, which necessitates a problem solving approach and the application of strategies." Tirkkonen-Condit (2005) makes similar observations, distinguishing a default translation procedure (or automaton) and a monitoring mechanism. The default translation automaton "operates on a lexical as well as syntactic level" (Tirkkonen-Condit 2005, 405) and produces translation hypothesis while the "monitor supervises text production processes, and triggers disintegration of the translation activity into chunks of sequential reading and writing behavior" (Carl and Dragsted 2012, 127). Also M&A (2022) classify keystroke pauses in (at least) two different bins, as unintentional halts and intentional pauses depending on their relative translator-dependent length. Their approach will be discussed in detail in Sections 3.2 and 5. A default-interventionist view on the translation process is also stipulated by the Monitor Model (Schaeffer & Carl 2013) in which automated translation routines form the basis of translation production that are monitored by higher-level cognitive processes.

### 2.4 Translation and affect

Dijksterhuis and Nordgren (2006) associate intuitive/automated Type 1 processes with unconscious thought. They characterize Type 1 as "task-relevant cognitive or affective thought processes" which includes choice of words and syntax. Conscious reflection and/or problem-solving Type 2 processes "cannot take place without unconscious processes being active at the same time." (ibid.) Type 1 processes, they say, "have to be active in order for one to speak" (ibid. 96). For Evans and Stanovich (2013) Type 1 processes are associated with basic emotions, they are "often manifest as feelings" (Evans 2010, 313) while deliberate Type 2 processing has bonds with more complex emotions that require deeper reflection and contextually appropriate reactions. However, Evans (2010) explains that "intervention on intuitions by reasoning requires both the cognitive capacity for the relevant reasoning and the awareness of the need for doing so" (Evans 2010, 323), suggesting that, in fact, there are three different processes in operation: intuition (Type 1), reasoning (Type 2), but also a process that brings intuition into conscious awareness. For Robinson (2023) this process of "becoming-cognitively" is a function of an "emotional interpretant". Drawing on Pierce (1866/1992), Robinson makes a distinction between an energetic, a logical and an emotional interpretant. Robinson (2023, 38) says: "Whatever conscious awareness we have of semantic meaning is constantly being fed affectively, conatively, and becoming-cognitively to the logical interpretant by the emotional and energetic interpretants." While the actual reasoning may then override or adjust the initial intuitions, it is first of all the awareness of phenomenal experience and emotions that trigger the need to intervene. I will come back to this discussion in the conclusion, Section 7.

Emotions influence how we interpret stimuli, guide attention and memory and alter our experiences. Emotions form the richness and shape our subjective awareness and perception of events. In this way, emotions serve as a lens through which we experience and understand the world, making our experiences more vivid and impactful. Hubscher-Davidson (2017) maintains that awareness of emotion plays a crucial role in the decision-making process of translators. According to her, translators often draw on emotional intelligence to understand the tone, mood, and context of the ST, which helps them convey the intended meaning accurately. She stresses that "Inevitably, translators will need to tap into their own resources of emotional experiences and emotional language in order to understand and transfer the information to the best of their abilities" (Hubscher-Davidson 2017, 4). Emotional engagement with the text can shape how translators deal with ambiguity or complex linguistic nuances.

Emotions can also affect cognitive load, impacting whether translators rely on automated/intuitive or more analytical processes during translation. But while emotions are internal to an agent, they surface though body language, mimetics, and nonverbal communication. Emotions, Robinson says, have an important social status: they are "performative responses to other people", they are "the glue that makes the world we cocreate with our environments cohere" and "afford us the ability to understand other people's feelings and feeling-saturated thoughts" (Robinson 2023, 86).

Given the generative assumption of human-environment interaction, it may thus appear unsurprising to assume that translators' keystroke and gazing behaviors can reveal their emotional and cognitive/logical states of mind, and that - by analyzing behavioral data gathered during translation sessions -, researchers may be able to approximate these assumed underlying mental states. In the next sections I will discuss and analyze two taxonomies that may be suited to provide criteria to detect automated, reflective and affective states of mind.

## 3 Fragmenting behavioral translation data

The temporal structure by which translations are produced has been a central topic of investigation in TPR since its beginnings in the 1980s (e.g., Königs, 1987), but with the usage of Personal Computers and the possibilities of keyloggers such as Translog (Jakobsen and Schou 1999) and InputLog (Leijten & Van Waes 2013), TPR has become technologically much more involved. While it is uncontroversial that translators sometimes engage in rapid, automated processing and sometimes demonstrate more deliberate, analytical behaviors, the analysis of IKIs provides the most accessible and measurable traces of these processes and thus open possibilities to accurately assess the mental processes by which translators presumably produce their texts. In addition, keylogging data can be complemented with eye-tracking data (Carl 2012, Carl et al 2016, Hvelplund 2016) which would then reveal not only how texts are produced, but also what information is taken in by the translator during the translation process.

In this section I suggest segmentation methods of behavioral translation data that account for the three ABC layers. The HOF taxonomy captures three different emotional states, mainly based on gaze data, while the TSF defines thresholds of IKIs to distinguish routinized/automated from non-automated/reflective processes during translation.

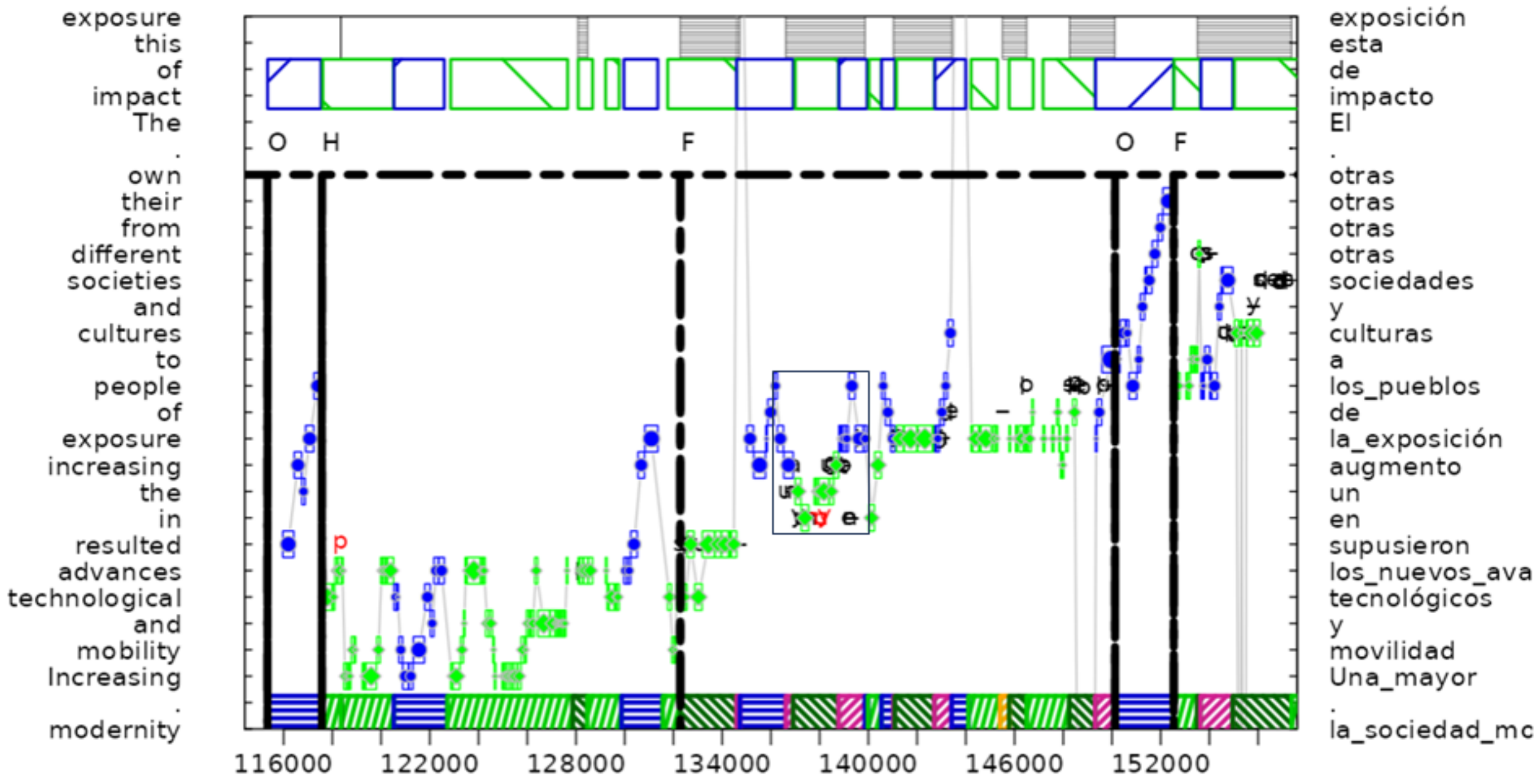

Figure 1 A progression graph of a small snippet of the translation session (BML12/P03 T5). The graph presents a segment of approximately 40 seconds (116,000 -156,000 milliseconds) of an English-to-Spanish translation. The vertical axis plots to the ST on the left side and the TT on the right side. ST and TT words are aligned on a word or phrase level. A single ST word that maps into TT phrase is marked as a multi-word unit on the right vertical axis, such as "Increasing → Una_mayor". An ST multi-word phrase that is translated into a single word appear as TT repetitions, such as "different from their own → otras otras otras otras". Blue dots and green diamonds indicate eye movements on the ST and TT respectively. The black and red characters are insertion and deletion respectively. Activity Units (AUs) fragment behavioral translation data into six categories, as marked in colored boxes at the bottom of the graph. The type (color) of the AU determines whether the translator is involved in reading the ST or TT, translation production, or simultaneously reading and writing (see Table 1). TSs are marked as gray boxes in the top of the graph. The rectangular box in the middle of the graph (around time 136,000 – 140,000) is reproduced in Figure 2 with a finer-grained segmentation of the TS into six (sub) Tasks.

### 3.1 The HOF Taxonomy

Carl et al (2024) developed a taxonomy that describes three basic affective/emotional translation states: a state of Hesitation, a state of Orientation, and a Flow state, the HOF taxonomy. An example showing behavioral data of these HOF states is plotted in the progression graph in Figure 1; a description is as follows:

- A state of hesitation (H) is characterized by surprise, when expectations do not fit the observation and may trigger re-reading, search, or repeated text modifications, etc. This state of mind indicates moments of uncertainty or cognitive challenge, signifying areas of complexity or struggles to find suitable translations in which a translator may be caught between choices or actions. Unexpected challenges may prompt the translator to revise or re-read a piece of text. Translators may feel a kind of internal conflict associated with momentary typing pause.
- A state of orientation (O) refers to the translator's behavior when feeling the need to get acquainted with the ST. It reflects a situation in which a translator aims at understanding the ST. It is a state of mind characterized by a longer stretch of ST reading. It may evoke specific cognitive responses that can lead to further search or reflection.

- A Flow state (F) is a phase in which the translator is immersed in translation production, generating the TT with ease and minimal interruption, and the individual becomes fully absorbed in the activity. It is marked by fluent translation production with minimal reading ahead and short pauses. Flow states have been described as an experience in which the agent loses track of time, the sense of self fades, and a seamless merging of action and awareness. Flow states involve optimal information processing, where a translator engages in undisturbed fluent translation.

Figure 1 shows a progression graph comprised of a sequence of five annotated HOF states (**OHFOF**) above the black dashed lines. Each HOF state consists of a sequence of Activity Units (AUs, Schaeffer et al., 2016; Hvelplund, 2016). AUs provide a fine-grained view on the translation process, focusing on the coordination of the translator's eyes and hands. AUs fragment behavioral translation data into six categories - marked in colored boxes at the bottom of the graph - depending on whether the translator reads the ST or the TT and/or engages in writing (see Table 1). For instance, the first Orientation Phase (marked **O**) in Figure 1 consists of one AU (ST reading), while the successive state of Hesitation (**H**) encompasses a sequence of 7 AUs in which the translators eyes switch from ST to TT reading, intercepted with concurrent typing .

| AU type | Reading / Writing Activity | AU Color in Figures 1 and 2 |
|---|---|---|
| T1 | ST reading | Blue |
| T2 | TT reading | Light Green |
| T4 | TT production | Yellow (no occurrence in the graphs) |
| T5 | ST reading with concurrent production | Red |
| T6 | TT reading with concurrent production | Dark Green |
| T8 | No observed behavioral data for more than one second | Black (no occurrence in the graphs) |

**Table 1:** : Types of AUs and color code in Figures 1 and 2.

### 3.2 The Task Segment Framework (TSF)

In their TSF, M&A suggest classifying IKIs into several categories:[5]

- Delays are IKIs < 200ms. Delays delimit *motor programs* which are the most basic typing patterns. According to M&A motor programs consist of 3-4 keypresses but several motor programs can cluster into one Task.
- Respites (*RSPs*) interrupt Tasks. A Task, in their terminology, is a sequence of motor programs, delimited by an *RSP*. An *RSP* indicates a "non-intentional" typing halt in which "attention and/or resources being drawn away from typing" (M&A, 26). Nevertheless, *RSPs* are considered "part of typing" (M&A, 23) as they are not intentional, for instance they do not involve planning. M&A define an $RSP$ to be $2 *$ median $within{-}word\ IKI$ (see Section 5.2)
- A *TSP* (Task Segment Pause) in contrast, is a consciously intended typing break.[6] *TSPs* delimit TSs, that is sequences of one or more Task(s). A *TSP* is thus longer than *RSP* and disrupts the typing flow. *TSPs* are "task-related" and "tend to happen between words and higher language units" (ibid. 23). M&A define a $TSP$ to be $3*$median $between{-}word\ IKI$ (see Section 5.2)
- M&A also mention the existence of "Superpauses" for instance to "accommodate first-pass readings" (M&A, 26). They define Superpauses to be much longer than *TSPs*, but do not address this type of pause in detail. I suggest that these Superpauses may be explained in the light of the HOF taxonomy as a state of Orientation on the phenomenal layer (see Figure 1).

[5] Here, I follow their suggestion that other researchers take up this framework because "applying the TSF may render different research projects on translation and other writing/typing tasks more comparable." (M&A, p. 26)
[6] There is no indication in M&A how "willful breaks" could be distinguished from involuntary breaks in process data other than by their duration.

Figure 2 shows a zoomed-in TS, which is an enlarged version of the small square in the center of Figure 1, that is part of a Flow state. The TS in Figure 2 consists of two Tasks separated by an *RSP*. Each Tasks can be characterized by several features, including its duration, the number of keystrokes, deletions and/or insertions, but also its gaze pattern (linear reading, re-fixation, scattered gaze path).

In the next section I address some properties of Tasks and TSs as gathered in a large corpus of translation behavioral data.

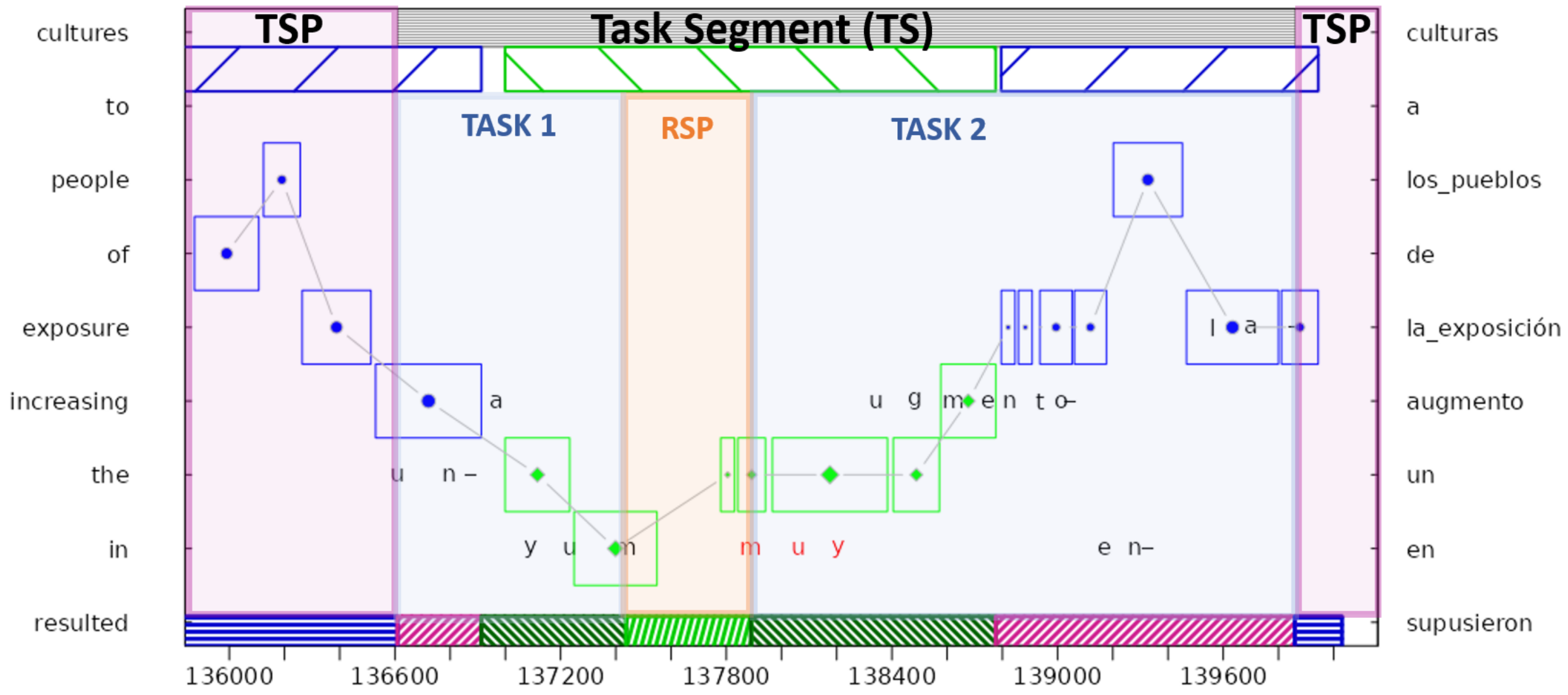

**Figure 2**. This Progression graph shows a segment of approximately 4 seconds (136,000 – 140,000) in which an English segment “in the increasing” is translated into Spanish “un augmento en la”. The duration of a TS is indicated as a grey bar at the top, which is preceded and followed by a *TSP* (violet boxes). The TS consists of two successive Tasks which are separated by an *RSP*. Each Task contains one or more insertion and/or deletion keystroke(s). In Task 1 the translator produces “un a yum”. The last tree letters “yum” are presumably a typo, because these letters are successively deleted again in Task 2 (deletion “muy” is inverse order of “yum”). Task 2 then shows the correct continuation “ugmento” (to produce “augmento”) and the production of “en la”. As indicated by the “Fixation Units”, i.e., the blue and green striped boxes on the top, while the eyes of the translator fixate in the beginning and the end of this segment at several ST words (blue boxes are ST fixations), they move to the target window during the correction or the typo (green boxes are TT fixations). The Sequence of colored AUs at the bottom of the graph indicates the coordination of reading and writing behaviour (see Table1). Note that Task 2 consists of two AUs of Type 6 and 5 which involve concurrent writing and reading of the TT and ST, respectively.

## 4 An empirical investigation

In Sections 4 5 and 6, I present an empirical analysis of the *TSF* and the HOF taxonomy based on a fragment of the CRITT TPR-DB (Carl 2016). In section 4, I present the data for the analysis, section 5 assesses the relations between Tasks and TS, while section 6 addresses their relation to HOF states.

### 4.1 The Empirical Data

The CRITT TPR-DB is publicly available under a Creative Commons License (CC BY-NC-SA), hosted at *sourceforge.net* and with ample documentation on the CRITT webpage[7]. It contains more than 5000 translation sessions and more than 600 hours of text production with recorded keystroke logging, mostly written translation sessions, but also authoring and spoken translation (sight translation, reading aloud, etc). In addition to the keystrokes, many studies have also gaze data.

The CRITT TPR-DB processes the raw logging data into several summary Tables (currently 11 different Tables per session with more than 300 Features) which describe the data from different angles. In this study we investigate the KD tables which provide information about the how keystrokes are produced, their production time, TT word and segment number to which the keystroke contributes, etc.[8]

[7] See https://sites.google.com/site/centretranslationinnovation/tpr-db/getting-started.

The data used in this study is part of the MultiLing sub-corpus which is a subset of the CRITT TPR-DB. The MultiLing corpus consists of six short English STs, each text has between 110 to 160 English words. The six STs have 847 words in total. Four of the six texts are news texts and two are excerpts from a sociological encyclopedia. These texts have been translated into several languages under different translation modes (from-scratch translation, post-editing, sight translation, translation dictation, etc). To make the translation processes better comparable, no consultation of external resources was permitted during the sessions.

This study only uses from-scratch translations that were translated using Translog-II (Carl 2012) ; only in section 4.4 I compare IKI distributions of from-scratch translation with post-editing. Section 4.2 compares IKIs of MultiLing texts into five languages (English to Arabic, Danish, German, Spanish, Hindi). Given that the Spanish and Arabic datasets show quite different IKI distributions (see Table 2 and Figure 3), I only look at those two target languages in the other sections.

Table 2 shows the TPR-DB (internal) study name, the target language, the number of total keystrokes in the study, the total duration of all sessions in hours, the number of sessions and number of different translators as well as the mean and median time in ms per IKI.[9] Table **2** shows that Danish translators are the fastest in our dataset: 24 Danish translators needed a total of 7.7 hours to produce 69 translations with 72.383 keystrokes and 383 ms per keystroke on average. Hindi translators, in contrast, were the slowest in this dataset; they needed on average 1223 ms per keystroke, more than 3 times longer.

| Study name | AR20 | BML12 | KTHJ08 | NJ12 | SG12 |
|---|---|---|---|---|---|
| Target language | ar | es | da | hi | de |
| #Keystrokes | 37171 | 73619 | 72383 | 43137 | 58883 |
| Total Duration (h) | 8.72 | 10.10 | 7.70 | 14.67 | 12.46 |
| #Sessions | 40 | 60 | 69 | 38 | 47 |
| #Translators | 22 | 32 | 24 | 20 | 24 |
| Mean IKI. in ms (log ms) | 844 (6.7) | 493 (6.2) | 382 (5.9) | 1223 (7.1) | 761 (6.6) |
| Median IKI, in ms (log ms) | 265 (5.6) | 156 (5.0) | 160 (5.1) | 374 (5.9) | 156 (5.0) |

**Table 2:** Properties of the empirical data used in this study. The study e.g., AR20 is the internal name in the CRITT TPR-DB, and has no further meaning in this article.

The English-to-Arabic translations were produced in the context of a PhD thesis (Ogawa et al 2021, Almazroei, 2025). The data was collected from 22 experienced PhD students at Kent State University. While this data consists of from-scratch translation, post-editing and sight translation, we only use the from-scratch translations in this study, with the exception of section 4.4, where also the post-editing data. The English-to-Spanish data was collected in 2012 from native Spanish translation student. This data been used since then in numerous studies (e.g., Mesa-Lao 2014, Schaeffer er al 2016, Gilbert et al 2023) and is described there in more detail.

[8] A full description of the Features in the CRITT TPR-DB is provided on the CRITT website: https://sites.google.com/site/centre-translationinnovation/tpr-db/features

9 Note that a keystroke can produce more than one character in the text, for instance in case of a deletion when a string was marked, or through copy & paste.

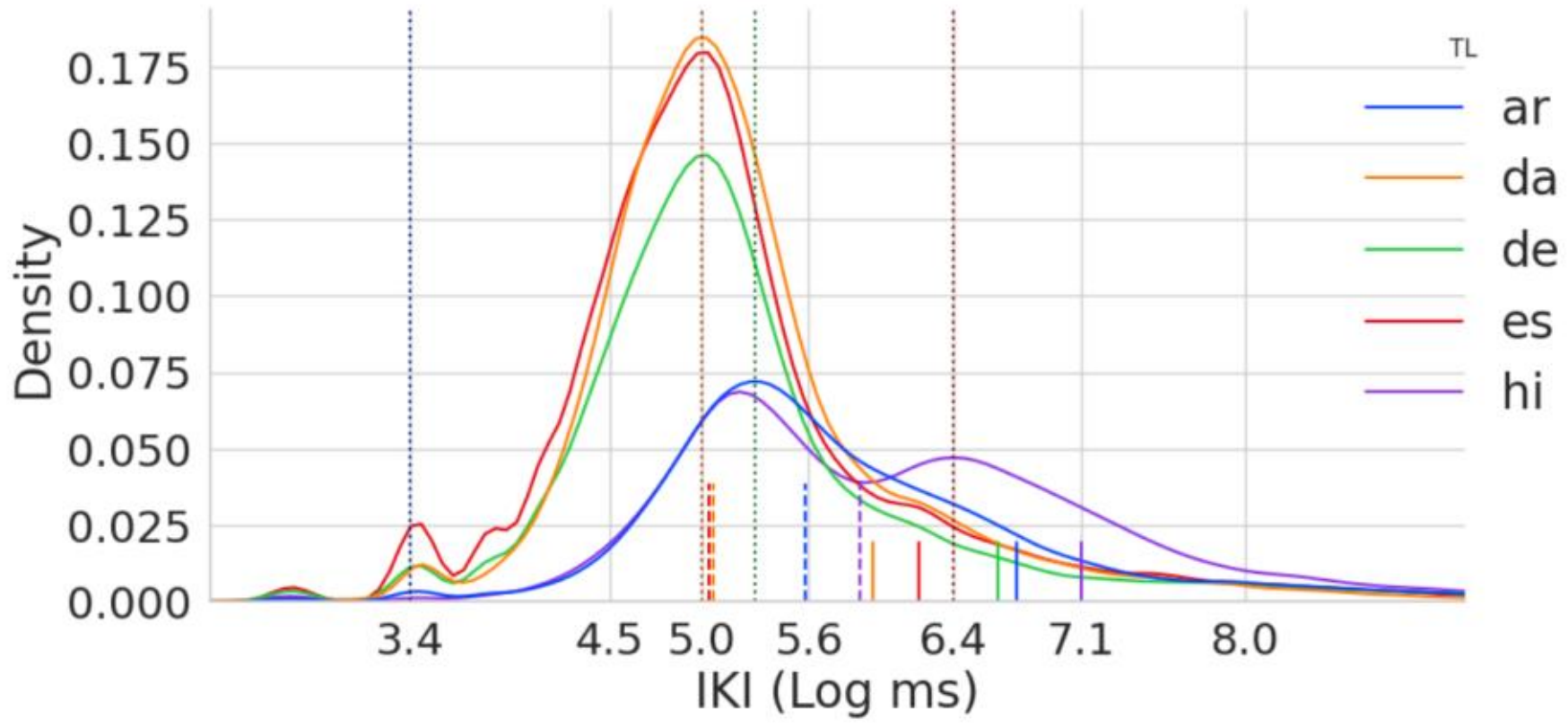

**Figure 3** Density of log transformed IKI distributions for five languages.

### 4.2 Distribution of IKIs across Languages

IKI distributions are heavily right-skewed. As suggested by Baaijen et al (2012) an IKI analysis may therefore be better suited based on the log-transformed data, which makes the distribution similar to a normal distribution.

Figure 3 shows the log-transformed IKI distribution of from-scratch translations from English into 5 languages (see Table 2). As can be seen, two language pairs (en→es and en→da) have a quite similar distribution with one (main) peak around 150 ms (5.0 on the log-transformed scale in Figure 3), whereas two language pairs (en→ar and en→hi) have a much flatter distribution with an IKI peak around 200 ms (marked as a vertical dotted line, but no numbered index on the x axis in Figure 3). There is an additional peak for Spanish (es), Danish (da) and German (de) around 30 ms (3.4 on the log-scaled x) and for the Hindi (hi) data around 600 ms (6.4). The mean and median IKIs are shown in Table 2 and marked in Figure 1 as shorter solid and dotted lines respectively. These mean and median values are computed from the original non-log transformed distributions (i.e., log(mean[X], rather than mean[log(X)]).

In their keystroke analysis of authored text, Baaijen et al (2012) make out two (or sometimes three) lognormal distributions which they attribute to within-word and between-word pauses, as well as pauses between sentences and subsentences. The average of within-word IKIs in their data, they say, are around 90 ms (4.5) within words, 270 ms (4.6) between words, 1212 ms (7.1) between subsentences and 2981 ms (8.0) between sentences. Some of these values are indicated as long vertical dotted lines in Figure 4, to better compare it with our data. IKIs below 30 ms (3.4) may be attributed to noise and can be ignored[10].

The log-transformed IKI distribution of our data is somewhat different from that of Baaijen et al (2012): it does not seem to carry indicators of two (or three) normal distributions (apart probably from the Hindi data) and the distribution of our peaks are different. As I discuss in Section 5, indicators for within-word (*WW*) and between-word (*BW*) IKIs in our data are quite different from those reported in Baaijen et al (2012). For instance, (as shown in Table 4 in Section 5) the mean IKIs in the 60 Spanish translation sessions is 697 ms for between word (*BW*) and 301ms for within word (*WW*) IKI.[11] Thus, what Baaijen et al (2012) report to be *BW*-IKIs (270 ms), we find in our Spanish data to amount approximately to median *WW*-IKIs (281 ms). Our findings are in many cases more similar to those of M&A who report median *WW*-IKI at around 330 ms, and median *BW*-IKIs around 735 ms.

### 4.3 Variation of Typing Speed

M&A cite (Van Waes et al. 2016) who allegedly "showed that 75% of all IKIs were below 250 ms" (M&A, 15). While this value relates presumably to writing (authoring), we can confirm a similar

10 We currently have no good explanation for the cause of the small IKI peak around 30 ms in our data.

11 Note these values are slightly different as compared to those reported in (Carl 2024) due to a modification how *WW* and *BW* are computed. We'll provide a more detailed discussion in Carl et al (forthcoming) value for our Spanish data (en→es). However, there is a considerable variation between different translators. Some of the Spanish translators produce merely 55% of the IKIs below 250 ms.

The typing speed also seems to depend on the TL. The average IKI duration for our Arabic translators is quite different from that of our Spanish translators: as Figure 4 shows, for en→ar there are on average only roughly 45% of IKIs below 250 ms; 75% of IKIs are below 550ms. Some Arabic translators need an average of 800 ms to produce 75% of the IKIs. There is currently no good explanation why the production speed of Arabic and Spanish translations are so different, given that both populations, the Arabic and the Spanish translators, were L1 native language advanced translation students.

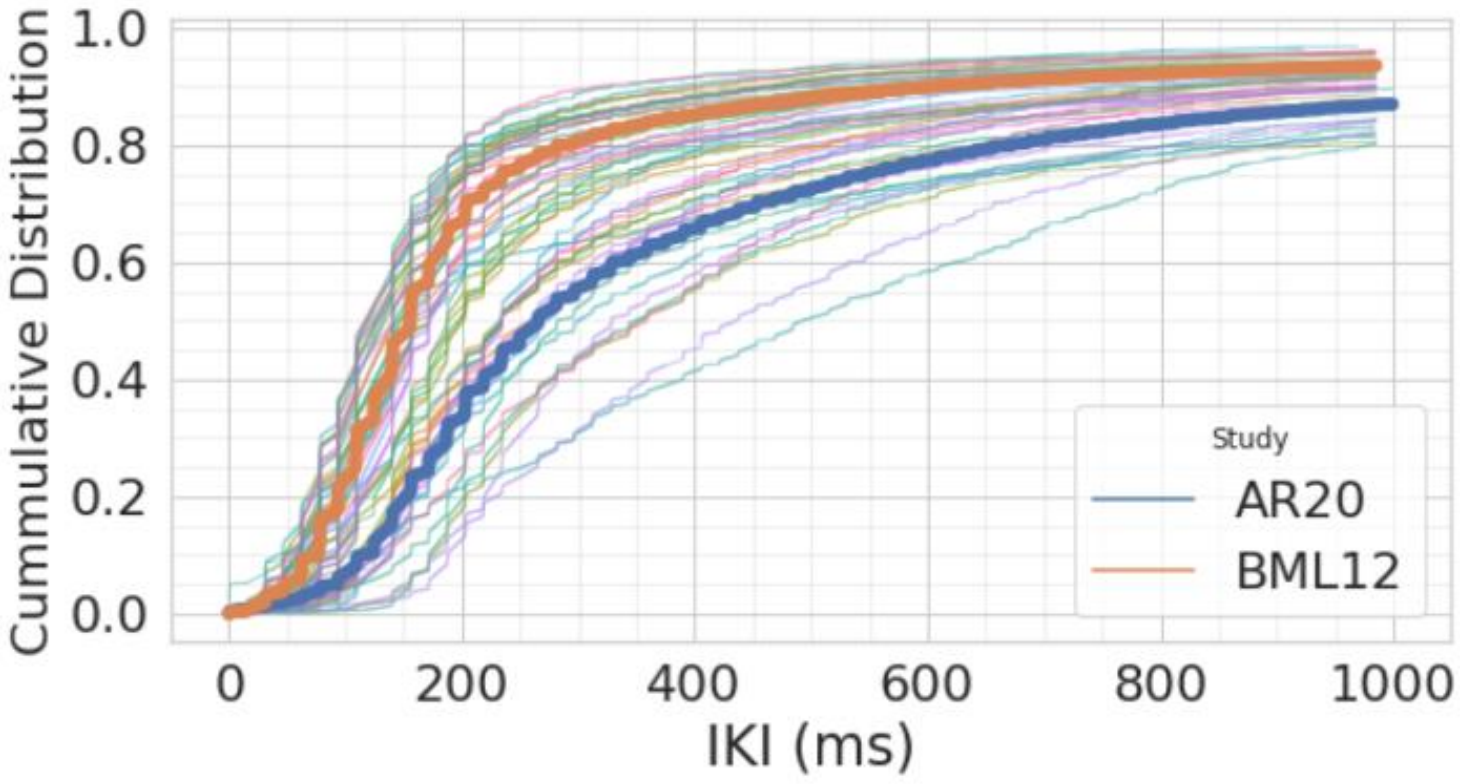

**Figure 4** Cumulative distribution function of IKIs for 32 Spanish and 22 Arabic translators as well as their averages (in bold). The graph plots different IKI profiles for the 54 different translators.

### 4.4 Inter and Intra-Translator Typing Speed

M&A (p. 12) also cite (Conijn, Roeser & van Zaanen 2019) who state that “overall IKI means are stable across tasks (copying, email writing, academic summarizing)”. M&A accordingly assume that “IKI values for smaller units may safely be assumed to be similar enough in regular typing and in typewritten multilectal mediated communication tasks” (ibid. 12). This statement coincides with Carl and Dragsted (2012) who find that typing speed in “unchallenged” (i.e., default) translation is similar to text copying.

In order to re-assess this hypothesis, we deployed the 2 sample Kolmogorov Smirnov Test (KS2). The KS2 test assesses how likely two samples are part of the same reference population and returns a p-value. The samples were considered to be from the same population (i.e., the same translator) if KS2 returns $p < 0.05$, otherwise the two samples would be considered different.

Many studies of the CRITT TPR-DB contain various translation and post-editing sessions from the same translators. For this experiment, we compared from the Spanish and Arabic data:

a) IKI distributions of from-scratch translations from 46 different translators, where each translator produced translations for two different texts.

b) IKI distributions of a from-scratch translation and IKI distributions of a post-editing session from 54 translators, where each translator translated and post-edited a different text.[12]

c) 1422 IKI distributions of two from-scratch translations from two different translators.

Table 3 shows that in 78% of the cases two translations of the same translator were indeed (correctly) recognized as samples from the same population (that is, the same translator). However, this was only the case for 44% of the translation and post-editing sessions produced by the same translator. In 96% of the cases IKI distributions from different translators were (correctly) classified as different.

[12] There are naturally much fewer keystrokes in post-editing than in from-scratch translation and IKIs between words are presumably much longer than IKIs within word. In order to compare IKI structure between from-scratch translation and post-editing, we only considered within-word keystrokes in this experiment. For the definition see section 5.1.

| | total | same # | same % | different # | different % |
|---|---|---|---|---|---|
| a) | 46 | **36** | **78** | 10 | 22 |
| b) | 54 | **24** | **44** | 30 | 56 |
| c) | 1422 | 55 | 4 | **1367** | **96** |

**Table 3** results of the KS2 test indication the number (#) and percentage (%) of same and different samples. Correct results are in bold.

This KS2 method is thus not suited to determine whether IKI distributions from different typing modes (translation and post-editing) were produced by the same typist. The 44% correct vs. 56% incorrect classifications in the b) experiment are just random guesses. However, the method can determine above chance (78%) whether IKI values of two different from-scratch translation sessions were indeed produced by the same translator.

## 5 The TSF: Pauses and Segments

It is generally assumed that translators mentally chunk the ST into portions that they translate as coherent typing burst (Malmkjaer 1998). Within TPR it is assumed that this mental chunking is reflected in the structure of IKIs. As pointed out in section 3.2, M&A define two IKIs thresholds that are considered here. *RSPs* and *TSPs* depend on the median within-words IKI (*WP*) and the median between-word IKI (*BP*) of individual translators, respectively. The distinction between *WP* (within-words) and *BP* (between-words) is well-established in TPR. It has been shown that *WPs* are shorter than *BPs* (e.g. Kumpulainen 2015), indicating different cognitive/mental activities at the boundaries of different levels of linguistic production. These translator-specific IKI thresholds may thus help separating those distinct mental processes. It is thus crucial to determine how to detect word boundaries in the flow of translation production.

### 5.1 Within-word and Between-word IKIs

In the keystroke logging data, every recorded keystroke is associated with a timestamp (Carl et al 2016). In this study, we define a *word boundary* to be any of the following keystrokes:

*Word-boundary keystrokes*: ` "'_.!?:=@$%&*()[]{}

, where blank spaces are mapped into an underscore '_'.

A keystroke is classified as *within-word* if it is not a word-boundary keystroke and it is neither preceded nor followed by a word-boundary keystroke. A *word-initial* keystroke is the first (non word-boundary) keystroke of a new word. Every IKI can then be classified according to whether it occurs within a word or whether it is word initial, and the pause before these keystrokes are WPs (within-word IKI) or BPs (before word IKI). A within-word IKI (*WP*) is preceded by a within-word keystroke, while the IKI preceding a word initial keystroke is defined to be the between-word IKI (*BP*). [13]

### 5.2 Respites and Task Segment Pauses

As translators have different typing skills and translation styles different *WPs* and *BPs* can be expected for every translator. Since the IKI data is heavily right skewed (see Figure 3) M&A suggest computing a median value, rather than a mean, as a basis for translator-relative pausing values. Following M&A, we define $RSP_i$ and $TSP_i$ for each translator *i* separately:

$$RSP_i = 2 * \text{median}(WP_i)$$

$$TSP_i = 3 * \text{median}(BP_i)$$

[13] In a forthcoming study (Carl et al forthcoming) we follow an easier definition, defining a *WP* as the IKI between two successive alphanumerical keystrokes and a *BP* as the IKI between a non-alphanumerical and a successive alphanumerical keystroke. The two methods produce slightly different but highly correlated *RSPs* and *TSPs* values.

Table 4 shows the summary information for *RSPs* and *TSPs* for the 32 Spanish and the 22 Arabic translators. As already discussed above, Table 4 shows that the various inter-keystroke pausing values for Arabic translators are much higher (almost twice) than those for Spanish.

| ar | min | max | mean | median |
|---|---|---|---|---|
| *RSP* | 312 | 1032 | 563 | 546 |
| *TSP* | 795 | 2388 | 1288 | 1077 |
| es | | | | |
| *RSP* | 220 | 470 | 301 | 281 |
| *TSP* | 423 | 1686 | 697 | 609 |

**Table 4**: *RSP* and *TSP* values for Arabic and English data

The minimum *RSP* duration in our data is 220 ms in the Spanish data, just above the assumed value for a Delay (200 ms, see section 3.2), while the maximum *RSP* duration is 1032 ms (in the Arabic data). The minimum *TSP* duration is 423 ms and the maximum is 2388 ms. Note that this maximum is still below the assumed third peak in the IKI distribution (see section 4.2), which ought to be around 2697 ms.

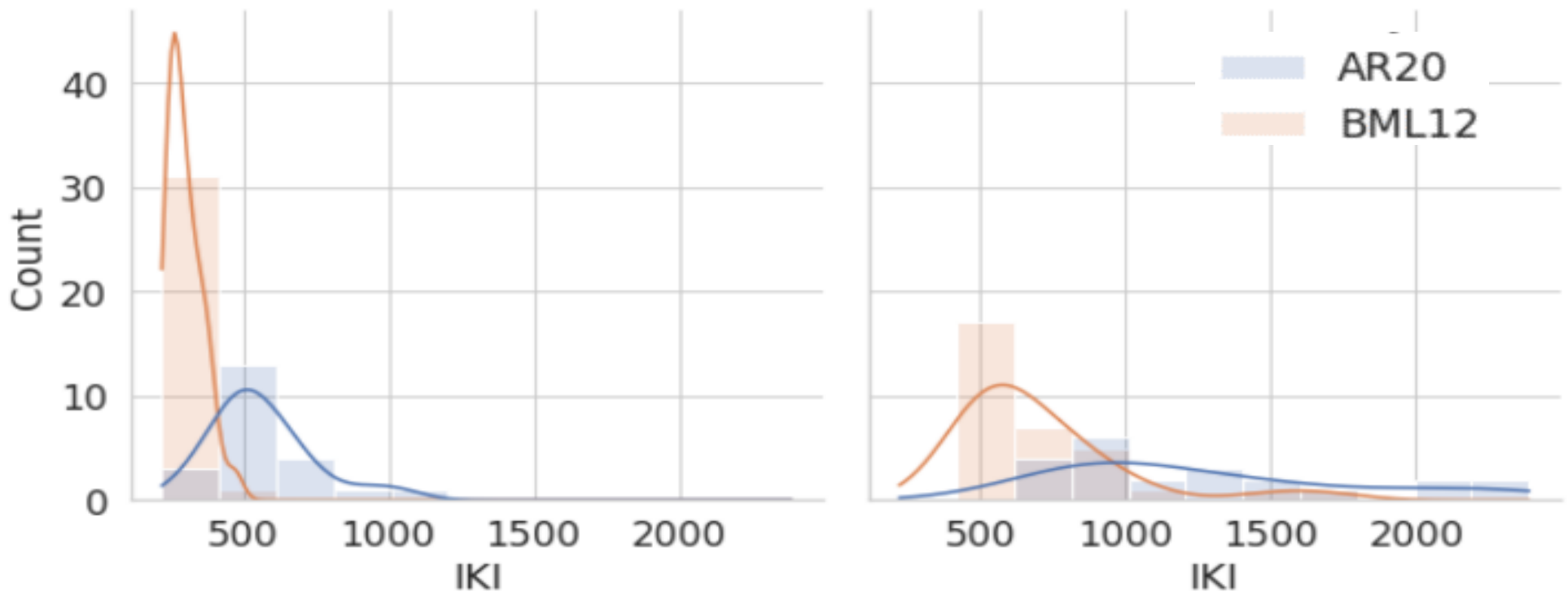

**Figure 5**: Distribution of RSPs (left) and TSPs (right) for Spanish and Arabic translators

### 5.3 Relating *RSPs* and *TSPs*

Figure 5 shows the distribution of *RSPs* on the left and *TSPs* on the right for 32 Spanish and 22 Arabic translators, respectively. As is the case for all IKIs (Figure 3), also *RSPs* and *TSPs* show larger variability for Arabic, i.e., a flatter distribution, than Spanish. However, all *RSPs* are - for every translator - shorter than their *TSPs*.

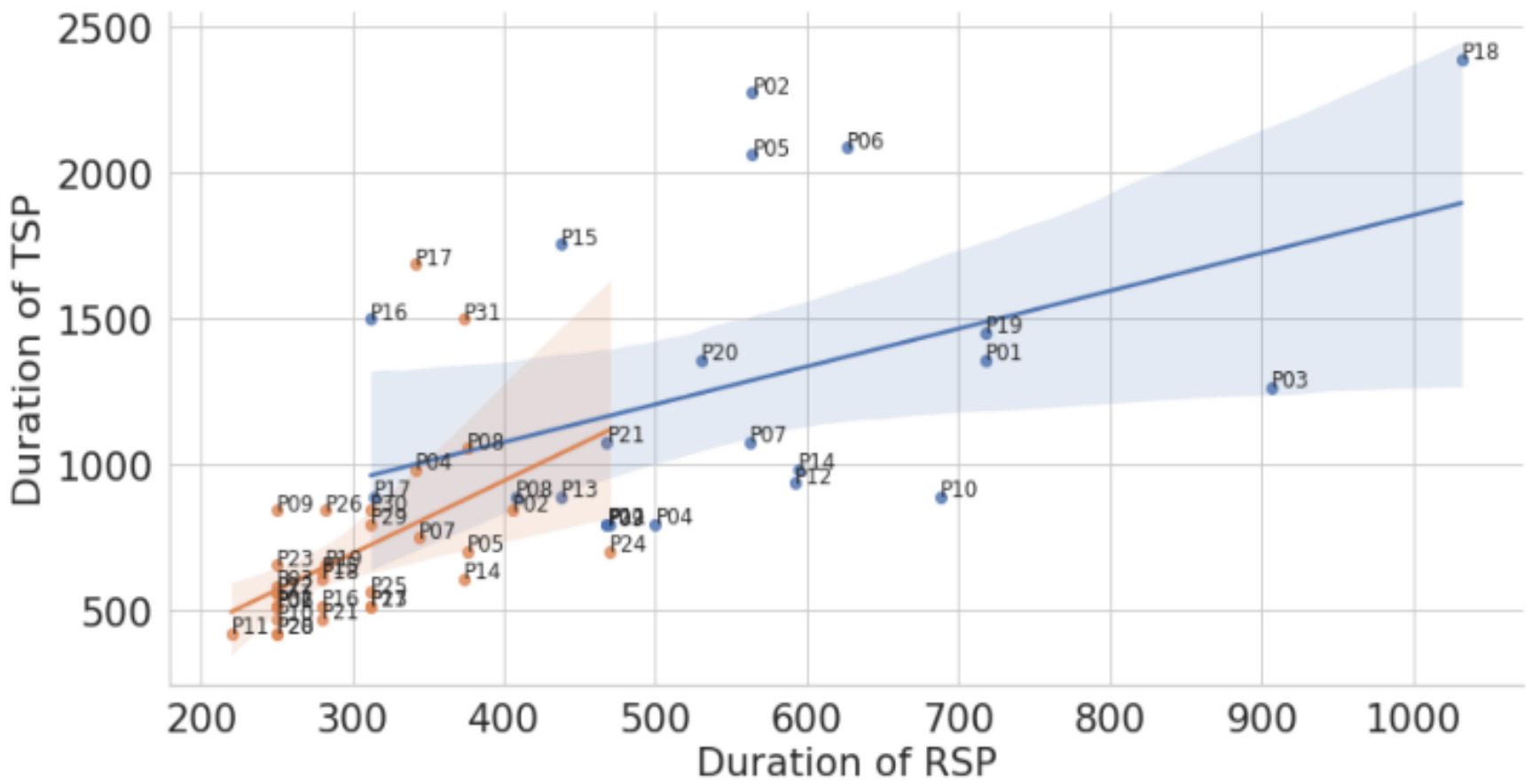

**Figure 6**: Correlation of *RSPs* and *TSPs*

Figure 6 shows that *RSPs* tend to correlate with *TSPs*. This correlation is significant for Spanish (Spearman τ:0.68, p < 0.0001), while it is not significant for Arabic (Spearman τ:0.40 p:0.065).

Interestingly, as plotted in Figure 7, there is a strong correlation between the number of Tasks (as delimited by *RSPs*) within a TS and the number of keystrokes produced in that TS (τ:0.74 and τ:0.73, p:0.000 for Arabic and Spanish respectively). While, on average, Arabic and Spanish translators engage in the same number of 2.2 Tasks per TS, Arabic translators show a larger variation (between min:1.2 and max:3.9, median:2.1) than Spanish translators (between min:2.1 and max:3.4, median 1.94).

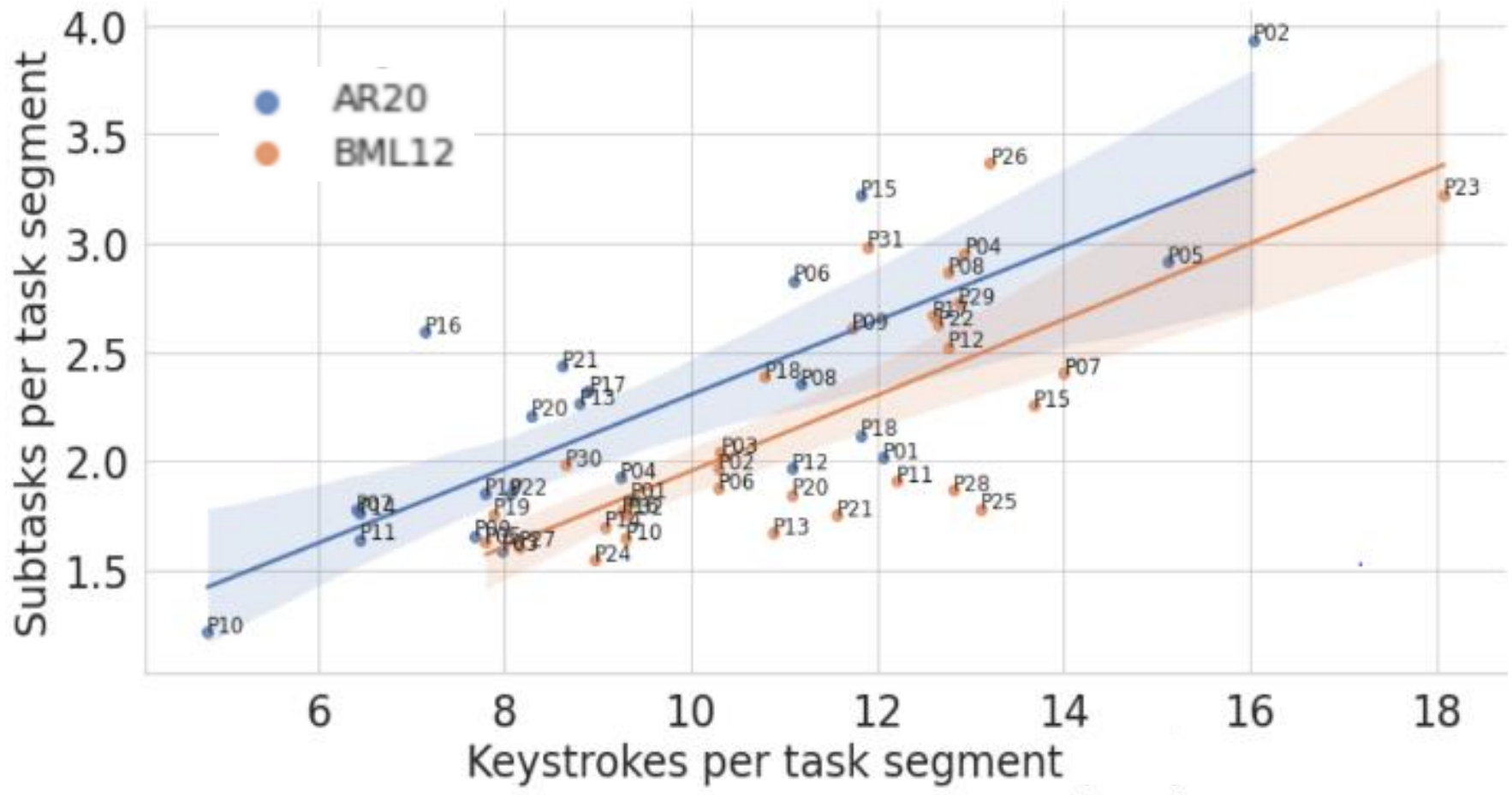

**Figure 7**: Correlation of total number of keystrokes per task segment and number of subtasks

Spanish translators also produce more keystrokes per TS than Arabic translators do. A Spanish TS contains between 8 to 18 keystrokes (mean 11.2) while an Arabic TS has between 5 to 16 keystrokes (mean 9.4). A Spanish Task has between 3.9 and 7.4 keystrokes (mean 5.3) while an Arabic Task has between 2.7 and 6.0 keystrokes (mean 4.3).

We also observe a slightly negative effect of TS length on the number of keystrokes produced per Task: As the number of  Tasks per TS increases, the number of keystrokes per Task decreases. This effect is significant for Spanish (τ:-0.52, p:0.002) but not for Arabic (τ:-0.18, p:0.41), which may have to do with the larger variability in the data and the smaller number of observations for our Arabic data set.

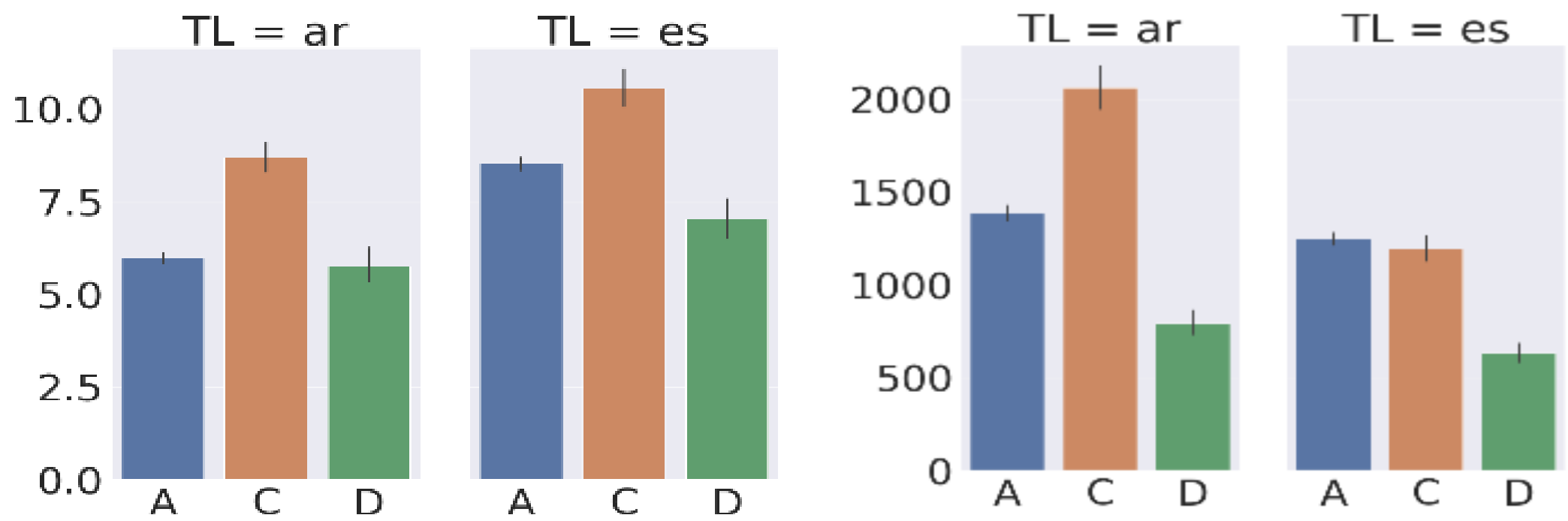

**Figure 8**: Number of keystrokes (left) and duration (right) for the three types of Arabic and Spanish Tasks. There are more keystrokes and shorter timespans in the Spanish data.

## 5.4 Types of Tasks

Following M&A, we distinguish between three types of Tasks that involve different types of keystrokes:[14] an insertion Task, **A**, has only insertion keystrokes (corresponds to M&A's ADD), a deletion Task, **D** (not considered in M&A) has only deletions, and a change Task, **C**, has insertions and deletions (corresponds to M&A's CHANGE). We omit the SEARCH Task, since in our translation sessions we have no external research.

[14] M&A suggest the following subtasks: ADD (adding new text, it was not clear to this author whether this is any insertion keystroke or only keystrokes at the end of the text), CHANGE (changing the text), and SEARCH (searching for information), HCI (human computer interaction), or it can consist of a combination of those. But the list seems to be open to further extension.

Figure 8 shows average duration and keystrokes for the three Tasks, for Spanish and Arabic. The figure shows that there are systematic differences between the Spanish and the Arabic tasks. It can be expected that differences exist also between individual translators, and presumably also for different text types and translation goals. We showed that IKI profiles seem to be typical for specific translators, so that translators can be recognized (to some extent) by their IKI distributions. Figure 8 shows that on average all types of Tasks **A**, **B**, and **C** have more keystrokes for Spanish as compared to Arabic and the average duration is longer for Arabic than for Spanish.

### 5.5 Types of Task Segments

A TS consists of sequences of Tasks, where each Task has a label (in our current taxonomy one of **A, C,** or **D**). We consider the sequence of Tasks labels realized within a TS to characterize the type of TS. Table 5 gives a summary of the 11 most frequent TS labels which make up 75% and 71% percent of Spanish and Arabic data respectively.

There are all together 10356 TSs in the Arabic and English data with 892 different TS labels. More than 93% of these TS labels, that is 833 different labels, occur less than 10 times. They account for 13.8% of the data (i.e., 1426 TSs). The 20 most frequent types of TS labels make up 90% of the data. The mean and median duration for all TSs is 6777 ms and 5781 ms respectively and the median, mean, and maximum number of Tasks per TS is 11, 14, and 61 respectively. In contrast, the mean and median duration for the 90% most frequent TSs, is 3479s and 3183s respectively and the number of Tasks is, on average, 3.0. Thus, most TS are relatively short, while there is a long tail of very long and diverse TS.

Table 5 provides labels of the 11 most frequent TSs, the total number of occurrences per TS, the percentage in Spanish and Arabic data, as well as their duration (in ms), the average IKI, and average number of keystrokes per Task. The most frequent TS, 38% and 36% of the Spanish and Arabic data, consists of a single Task **A**. There are on average 5.33 keystrokes for this Task with an average IKI of 173ms.

| TS Label | #Occur. | %Spanish | %Arabic | Dur. of TS | Average IKI | Key/Task |
|---|---|---|---|---|---|---|
| A | 3870 | 37.95 | 36.46 | 921 | 173 | 5.33 |
| AA | 1398 | 13.94 | 12.81 | 2167 | 211 | 5.13 |
| D | 753 | 7.71 | 6.59 | 504 | 121 | 4.16 |
| AAA | 543 | 5.49 | 4.86 | 2740 | 190 | 4.81 |
| AAAA | 263 | 2.72 | 2.25 | 4365 | 233 | 4.67 |
| DA | 194 | 2.15 | 1.44 | 1593 | 196 | 4.07 |
| AD | 164 | 1.35 | 1.96 | 1607 | 226 | 3.56 |
| C | 164 | 1.27 | 2.08 | 641 | 152 | 4.23 |
| DD | 116 | 1.42 | 0.64 | 1183 | 122 | 4.84 |
| AAAAA | 107 | 1.06 | 0.99 | 5300 | 230 | 4.61 |
| CC | 84 | 0.62 | 1.11 | 1181 | 160 | 3.70 |

**Table 5**: The 11 most frequent types of TS and their percentage for Spanish and Arabic. The column #Occur shows the total number of *TS*s and %Spanish and %Arabic the proportion in the two languages. Dur of TS provides the averages duration of the TS in ms. The table shows the Average IKI and average Keystrokes per Task (Key/Task).

As previously mentioned, the average number of keystrokes per Task decreases as the number of Tasks in the TS increases. There are 5.13 keystrokes per Task if the TS consists of two **A** Tasks, 4.81 keystrokes if the TS has three **A** Tasks, 4.67 for four Tasks, etc. On the other hand, the IKIs tend to increase as the TSs become longer, which suggests that typing becomes more interrupted.

Note the very strong correlation between frequencies (percentages) of the Spanish and Arabic TS labels (r=0.998). This indicates that Spanish and Arabic translators in our dataset engage in very similar production processes (i.e., sequences of Tasks) which results in very similar relative number of occurrences. This certainly needs verification in other datasets, but it might show a language and translator independent translation universal.

Olalla-Soler (2023) investigates successive ADD tasks within a TS, separated by *RSPs*. He defines *default translations* to be a sequence of fluent typing which consists of one or more ADD tasks that have, among other things, only a few *RSPs* (fewer than that of 75% of all TSs). Olalla-Soler (2023) observes that 67.8% of his TSs were ADD-only. These ADD-only segments contained 69.5% of the words. Our observations show that slightly more than 60% of the TSs are **A**-only and they cover around 44% of the keystrokes.

# 6 The HOF taxonomy

This section assesses a subset of the Spanish and Arabic data that was manual annotations with HOF states. It consists of eight Spanish sessions and six Arabic translation sessions, as described in detail in (Carl, Sheng, Al-Ramadan 2024). Table 6 provides an overview of the number of annotated states in the eight Spanish sessions and six Arabic sessions. Despite the different absolute numbers, it is interesting to note that the percentages of H, O and F states is almost identical in the two languages.

| | O | %O | F | %F | H | %H | Total |
|---|---|---|---|---|---|---|---|
| es | 183 | 30% | 284 | 47% | 139 | 23% | 606 |
| ar | 93 | 32% | 132 | 45% | 67 | 23% | 292 |

**Table 6**: Number and Percentages of HOF translation states in the annotated Spanish and Arabic data. There are approximately half the number of states for Arabic for 25% less annotated data.

## 6.1 HOF states

Table 7 shows a transition matrix between HOF states, for Arabic and Spanish data. The first row indicates the state at time *i* from where the transition starts, while the columns indicate the transition probability into the next state at time *i+1*. As can be seen, the most frequent pattern is a loop over Orientation (O) and Flow (F) states. Only in 16% and 14% of the cases for Arabic (ar) and Spanish (es) respectively, is an Orientation state followed by a Hesitation. However, both transition matrices are quite similar, with the only obvious exception that Arabic translators transition more often from a Hesitation to a successive Orientation (21%) while this is much more unlikely for Spanish translator (9% of the cases). In both cases, perhaps not surprisingly, the highest chances are that a translator will try to arrive at a Flow state (F).

| | | ar | | | es | | |
|---|---|---|---|---|---|---|---|
| | To | O | F | H | O | F | H |
| From | O | - | 0.84 | 0.16 | - | 0.86 | 0.14 |
| From | F | 0.60 | - | 0.40 | 0.60 | - | 0.40 |
| From | H | 0.21 | 0.79 | - | 0.09 | 0.91 | - |

**Table 7**: Transition Matrix between HOF states for Arabic (left and Spanish (right). Rows add up to 100%

## 6.2 Distribution of Tasks in H and F states

**Table 8** shows the distribution of **A**, **D**, and **C** Tasks in Hesitation (H) and Flow (F) states for the two languages.[15] According to this table, as can be expected, Flow states are clearly dominated by **A** Tasks while deletions and additions are more equally distributed during Hesitation.

[15] One assumption in this framework is that every Task is completed within one TS and every TS is completed within one HOF state. That is, Tasks are not supposed to overlap between two (or more) TSs and every TSs must be completed within one HOF state

Table 9 confirms the assumption that different TS patterns are realized in H and in F states. The table shows the six most frequent TS labels of F and H states in the two languages. This accounts for roughly 75% of the adjusted TSs. The table shows a very strong correlation (r=0.993, for the first 20 labels) between the Arabic and Spanish Flow states and between the Arabic and Spanish Hesitation states (r=0.968). The correlation between Flow and Hesitation states is slightly lower, r=0.85 and r=0.76 when changing both language and type of state.

| | ar | | es | |
|---|---|---|---|---|
| | H | F | H | F |
| **A** | 0.54 | 0.84 | 0.53 | 0.81 |
| **D** | 0.34 | 0.08 | 0.41 | 0.08 |
| **C** | 0.12 | 0.08 | 0.06 | 0.11 |

**Table 8**: Percentage of Tasks in Flow and Hesitation states for the Spanish and Arabic data. There is clearly a higher proportion of **A** Tasks in Flow states but proportionally more **D** Tasks during Hesitation. (Columns add up to 100%)

| F:ar | F:es | H:ar | H:es |
|---|---|---|---|
| **A** | **A** | **A** | **A** |
| **AA** | **AA** | **D** | **D** |
| **AAA** | **AAA** | **AA** | **C** |
| **AAAA** | **AAAA** | **C** | **AA** |
| **C** | **C** | **DD** | **DA** |
| **D** | **D** | **DA** | **CA** |

**Table 9:** Six most frequent TS labels for Flow in Hesitation states in the Arabic and Spanish. Note the identical ranking of TS frequencies in the Flow state.

### 6.3 Time structure of HOF states

Table 10 provides a summary of Flow state properties for the Arabic and Spanish translation sessions. The Table indicates the mean, minimum and maximum values for various features per Flow state. Table 10 shows that Arabic Flow states are approximately twice as long as the Spanish ones (Dur/F). There are more keystrokes per Arabic Flow states (Keys/F), there is a larger number of TSs (TS/F) and Tasks (Tasks/F) and the variation of these parameters is higher in Arabic data as compared with the Spanish data. However, the number of keystrokes per Task (Keys/Task) and the number of keystrokes per TS (Keys/TS) seem to be lower in Arabic Flow states than in Spanish Flow states. Consistent with the analyses before, this indicates that Arabic translators are slower, and their translation proceeds more disrupted.

A slightly different pattern can be observed for Hesitations as shown in Table 11. As for the Flow states, the duration of Hesitations is also longer for Arabic in comparison with Spanish and there are more keystrokes, Tasks, and TSs in Arabic than in Spanish Hesitation states. However, the number of keystrokes during a Hesitation state is significantly lower as compared to Flow states. There is a large difference in the number of keystrokes during Hesitations in the Arabic (mean 17.68) and the Spanish (mean 3.19), almost 500%. This large difference may indicate a less clear discrimination between states of Hesitation and Flow in the Arabic data before transitioning to the next HOF state. However, presumably due to the manual annotation, in 6.4% of the cases (666 of 10356) a TS crossed a HOF state boundary. Most common overlapping TS were observed in the transitions H → F, F → O, F → H and O → F with 136, 133, 132, and 131 occurrences respectively. These Task Segments were cut at the transition between HOF states, which resulted in more and slightly shorter average TSs.

| ar | Dur/F | Keys/F | TS/F | Tasks/F | Keys/TS | Keys/Task |
|---|---|---|---|---|---|---|
| mean | 12320 | 36.58 | 3.04 | 7.17 | 12.50 | 5.04 |
| min | 8314 | 23.48 | 1.48 | 5.22 | 10.67 | 4.10 |
| max | 21930 | 63.33 | 5.60 | 10.40 | 15.88 | 6.09 |
| es | | | | | | |
| mean | 6389 | 29.42 | 2.10 | 5.44 | 14.85 | 5.47 |
| min | 3585 | 19.52 | 1.30 | 4.00 | 10.76 | 4.69 |
| max | 9484 | 40.97 | 3.81 | 7.80 | 20.04 | 6.70 |

**Table 10**: Summary information of Flow states for the Arabic and Spanish data. Mean, minimum and maximum values for average duration (Dur/F), keystrokes (Keys/F), TSs (TS/F) and Tasks per Flow state (Tasks/F). The table also shows average keystrokes per TS (Keys/TS), and keystrokes per Task (Keys/Task) in Flow states.

.

| ar | Dur/H | Keys/H | TS/H | Tasks/H | Keys/TS | Keys/Task |
|---|---|---|---|---|---|---|
| mean | 14155 | 17.68 | 2.76 | 4.54 | 6.78 | 3.99 |
| min | 9703 | 11.15 | 2.14 | 3.62 | 3.82 | 2.51 |
| max | 17597 | 24.14 | 3.38 | 6.26 | 11.27 | 6.26 |
| es | | | | | | |
| mean | 7329 | 3.19 | 1.95 | 2.23 | 1.55 | 1.34 |
| min | 4199 | 1.71 | 1.53 | 1.55 | 1.07 | 1.02 |
| max | 13122 | 7.64 | 2.59 | 3.32 | 2.95 | 2.30 |

**Table 11**: Summary information of Arabic and Spanish Hesitation states. Same columns as in Table 10.

This assumption is corroborated by the fact that Spanish translators spend approximately 18% of their Flow states in longer keystroke pauses (*TSPs*), while the pausing time during Hesitations is 3 times higher, 54%. That is, 54% of the Hesitation time is (presumably) spent in reflection or visual search (no keystrokes produced), and 46% in the completion of Tasks.

Arabic translators, in contrast, spend 26% of their Flow-state time in deliberate pausing (*TSPs*) and less than twice that amount, 45%, during Hesitation. Thus, the discrimination between Flow and Hesitation seems fuzzier in the Arabic data as compared to the Spanish one. The boundary between reflection and typing, Flow and Hesitation seems to be less pronounced for the Arabic translators.

Finally, the slower (or more meticulous) pace of Arabic translators is also obvious in the duration of Orientation states (see Table 12), which - similar to the other HOF states - amounts to approximately twice that of their Spanish colleagues.

| | mean | min | max |
|---|---|---|---|
| ar | 10300 | 5851 | 15169 |
| es | 4838 | 3007 | 8050 |

**Table 12**: Duration of Orientation states for Arabic (ar) and Spanish (es) translators.

## 7 Discussion and conclusion

This article proposes a model of the human translating mind as a structure composed of three embedded and interacting processing layers, each addressing distinct mental functions: an affective layer realizing emotional states **(A)**, a behavioral layer executing automated processing routines **(B)**, and a cognitive layer enacting deliberate and reflective thought **(C)**. The model integrates two taxonomies: the HOF taxonomy categorizes three affective/emotional translation states, and the TSF identifies behavioral indicators of automated and reflective translation processes. Using these frameworks, the article analyzes translation behavior by examining recorded keystroke and gaze data, revealing traces of the underlying ABC processes.

Sequences of *motor programs* form the basis of human translation production. Following M&A, sequences of motor programs cluster into translation *Tasks* by which translators add, change, or delete a piece of translation. Sequences of translation Tasks are separated by short interruptions in the flow of keystrokes, so-called respites (*RSP*s). Sequences of Tasks, in turn, form Task Segments (TSs) which are separated by a slightly longer Task Segment Pause (*TSPs*). According to M&A, *RSPs* are involuntary distractions, while *TSPs* are intentional, willful breaks in the translation process. Tasks and TSs are thus indicative of automated/routinized processing (B-layer) while *TSPs* signal reflective thought or translation planning (C-layer).

The article shows that different translators follow different pausing patterns, and that different types of Tasks and TSs are characteristic for different phenomenal, affective/emotional translation states (A-layer). We observe large variations in the temporal structure of typing behavior between different translators, but each translator seems to produce largely consistent and recognizable temporal typing patterns, indicating stable individual translation styles. While the study shows that specific typing/pausing styles are typical for different translators and affective states, it may be expected that they also depend on text difficulty, translation goals, and translator expertise, among other things.

Despite the different temporal translation production patterns across different translators, the study also shows a strong similarity in the types of activities that translators engage in. While the typing speed may vary by more than 100% across the two languages, there is a very strong correlation in the *types* of Tasks that translators perform (adding, deleting, changing text), leading to almost identical frequency distributions of TS Labels in the Arabic and Spanish data. Thus, while the temporal structure (the "how", i.e., temporal structure of typing pauses) seems to be highly individual, the content (the "what", i.e., patterns of insertions and deletions) seems to be quite similar across the two languages investigated, and presumably across translators.

The variation in the temporal structure of translation production—how translation unfolds over time as a dynamic cognitive-affective and embodied process—relates closely to the ABC account of the translating mind. Temporal features such as pauses, restarts, and revision loops often correlate with the translator's embodied hesitation, emotional response, or a somatic dissonance. For instance, a hesitation may not only reflect cognitive uncertainty but affective resistance or embodied discomfort with a translation choice. Translation, then, is not just a rational, disembodied activity but is deeply shaped by bodily rhythms, emotional attunement, and affective resonances.

The suggested three-fold ABC architecture of the translating mind is reminiscent of Robinson's (1991, 2023) ideosomatic theory of translation. Robinson's ideosomatic theory posits that translation is not purely a cognitive or linguistic task but involves the translator's physical, emotional, and sensory experiences—what Robinson calls "somatic" responses. The ideosomatic theory highlights how translators mediate embodied experiences and social norms. The term "ideosomatic" combines *ideo*- (referring to ideas or beliefs) and *somatic* (relating to the body). Robinson suggests that translators not only work with words but also engage their bodies and feelings when interpreting and translating text. These somatic responses, influenced by the translator's cultural and individual backgrounds`, affect decisions and interpretations throughout the translation process. For instance, translators might feel that a certain translation "feels right" or "off," guiding them toward certain choices based on embodied responses rather than solely cognitive analysis. This study suggests that the temporal structure of these processes is highly individual, while the sequencing of affective and behavioral states seem to be similar across different translators.

Robinson (2023) borrows and extends Peirce's (1866/1992) semiotics, emphasizing subjective experience of translators. Following Peirce, Robinson posits three types of interpretants, *emotional*, *energetic*, and *logical*, which roughly correspond to the ABC layers of the translating mind. Each interpretant unfolds concurrently and represents different ways a sign (word, phrase or text) can be interpreted or understood:

A. **Emotional Interpretant**: implies an initial feeling to a text or sign, a feeling or emotional response which actively shapes interpretation, cognition, and thus translation. The emotional interpretant is the basis for different HOF states, flow, hesitation and orientation, which actively shape cognitive and motor aspects of translation

B. **Energetic Interpretant**: refers to a reaction that involves effort or action. It involves a physical or motor response to a sign, which is, in Robinson conception, inherently embodied. The energetic interpretant thus highlights the embodied/energetic nature of behavioral processes that directly interact with the environment.

C. **Logical Interpretant**: involves conceptual or cognitive processes; rational, or reflective conclusions, shaped by habits or general rules of reasoning. The logical interpretant allows for enduring conceptual insights, that, for Robinson, imply a cognitive-affective synthesis.

Robinson (2023, 81) stipulates that two expressions share the same sense, if they evoke a “shared kinesthetic sensation.” As we read a text, Robinson says, we nonverbally experience its (deverbalized) sense on an affective/emotional layer. This brings a translator into a relation with the text which affords him/her “the ability to understand other people’s feelings and feeling-saturated thoughts” (ibid. 86). Emotional responses can trigger intuitive judgments and the possibility to translate (for humans). The reflective mind of the logical interpretant can, in contrast, consciously regulate emotions, potentially mitigating or amplifying their impact on decision-making. A translator's emotional response might influence immediate intuitive decisions, but reflective reasoning can override or reinterpret those emotions for more deliberate translation choices.

Thus, a flow state of smooth translation production can be arrived at and maintained if the emotional interpretant is “in tune” with the actions and observations of the energetic interpretant. In case of a clash of predictions and observations between these two interpretants, the logical interpretant may be called upon to re-integrate sensorimotor contingencies of the energetic interpretant with expectations of the emotional interpretant. Robinson calls this the “feeling-becoming-thinking” which is a central part of his somatic theory of translational action. While feeling and thinking are subjective, Robinson (2003) maintains that these translator-internal processes (which, for him, are qualia) are also “transcranial”; they are observable from outside: “we do often have more or less reliable access to the qualia [i.e., the subjective experiences] in other people’s heads, with or without verbal reports.” Qualia, he says, are often transcranial—"intracraniality is itself mostly transcranial” (ibid. 26), shaped through social interactions and shared experiences. While communicated, among other things, via mimic and gesticulation, here we assume that they can also be traced in the temporal structure and coordination of gaze and keystrokes during translation.

Robinson’s ideosomatic translation theory suggests that effective translation relies on a dynamic interplay between emotional intuition and reflective reasoning. Translators must navigate between immediate, affect-driven insights and more controlled, logical evaluations to produce coherent and contextually appropriate translations. This study aims at disentangling this relationship by examining the temporal structure(s) of A: translation states, B: Task types and C: Task Segments that might provide answers to the three embedded layers of the (translating) mind. While the temporal structure is highly individual and varies significantly across translators, the study shows that the sequencing of Tasks and affective translation states remains remarkably consistent. This consistency points to shared cognitive processes underlying translation, despite stylistic differences in execution. Combined with these empirical findings, Robinson’s ideosomatic theory offers a nuanced framework for understanding these cognitive-emotional dynamics of translation, showing how individual styles emerge from common mental operations.

**Funding**: “This research received no external funding”

**Data Availability Statement**: The data used in this article are freely available and can be downloaded from the CRITT website. The CRITT provides free server access through registration via: **https://sites.google.com/site/centretranslationinnovation/tpr-db/getting-started** (accessed 22 Feb. 2025). Upon logging into the CRITT server as *summer_gst*, a Python notebook is available under *shared/IKI_analysis.ipynb* that contains the Python code.